\documentclass[letterpaper]{article}  
\usepackage{aaai2026}                 
\usepackage{times}                    
\usepackage{helvet}                   
\usepackage{courier}                  
\usepackage[hyphens]{url}             
\usepackage{graphicx}                 
\usepackage{natbib}                   
\usepackage{caption}                  
\usepackage{booktabs}                 
\usepackage{tikz}                     
\usetikzlibrary{positioning,arrows.meta}
\nocopyright
\copyrighttext{Camera-ready version. ICAPS 2026 Workshop on Hierarchical
Planning (HPlan 2026), Dublin, Ireland. Non-archival.}

\title{HTN Planning as a Coordination Layer for Multi-Server\\
       MCP Tool Orchestration}
\author{
    Eliott Jacopin\textsuperscript{\rm 1},
    \'Eric Jacopin\textsuperscript{\rm 2},
    Koichi Takahashi\textsuperscript{\rm 1}
}
\affiliations{
    \textsuperscript{\rm 1}RIKEN, Kobe, Japan\\
    \textsuperscript{\rm 2}Cosmic AI, France\\
    eliott.jacopin@riken.jp, eric.jacopin@protonmail.com, koichi.takahashi@riken.jp
}

\begin{document}

\maketitle

\begin{abstract}
The Model Context Protocol (MCP) isolates servers by design: only the host
can orchestrate cross-server workflows. When the host is a large language
model, the resulting orchestrations are non-deterministic, non-reproducible,
and pay one inference round-trip per tool call. We present a coordination
architecture in which a Hierarchical Task Network (HTN) planner generates a
verifiable cross-server plan once, and a runtime middleware executes it
deterministically across multiple MCP servers, binding cross-action data
dependencies via a template mechanism (\verb|${context.X}|) substituted at
execution time. The architecture mirrors MCP's isolation constraint: each
compound task decomposes into server-local primitive actions, and inter-server
data flow is bound at execution time via JSON-path output extractors. We
instantiate the architecture on five HTN domains spanning laboratory robotics,
bioinformatics and multiscale modelling, and demonstrate end-to-end execution
from a browser-based plan controller against eight live third-party MCP
servers querying real biological databases.
\end{abstract}


\section{Introduction}
\label{sec:intro}

The Model Context Protocol \citep{mcp2024spec} has rapidly become a
\emph{de facto} standard for exposing external tools, databases and APIs
to LLM-based agents. An MCP ecosystem of $n$ servers offers a combinatorial
space of tool sequences, but its servers are isolated by design: a server
cannot invoke a tool on another server. Cross-server workflows must therefore
be orchestrated by the host. The dominant practice is to let the LLM decide
each next tool call from the current observation, in the manner of
ReAct-style agents \citep{yao2023react,schick2023toolformer}, with
recent MCP-based scientific assistants following the same
pattern~\citep{ruscone2025prototyping}. For
exploratory interaction this works; for workflows that require
reproducibility---laboratory automation, scientific protocols, regulated
pipelines---it fails on three counts. First, the same goal may produce
different tool sequences across runs. Second, an $N$-step workflow incurs
$N{+}1$ LLM inferences, each adding latency and cost. Third, no plan
exists prior to execution that a domain expert can inspect and authorise.

A formal planner sidesteps all three problems, and Hierarchical Task
Network planning is a natural fit: methods encode the procedural
decompositions domain experts already carry in their heads, and primitive
actions can be mapped directly onto MCP tool invocations. Authoring the
HTN domain from a collection of MCP servers is itself a non-trivial
problem; we treat it separately in a companion paper
\citep{anonymous2026authoring}, which covers the multi-pass LLM
compilation pipeline that produces the artifacts this paper consumes.
The present paper concerns the downstream coordination problem.

\paragraph{Research question.} \emph{Given an HTN domain over an MCP tool
inventory, can a planner-plus-middleware architecture coordinate
cross-server execution deterministically and verifiably, and how does this
compare with LLM-driven tool orchestration in cost and reproducibility?}

\paragraph{Contribution.} We describe a three-layer coordination
architecture---planning, binding, execution---in which a GTPyhop planner
\citep{nau2021gtpyhop} produces a sequence of primitive actions over an
MCP tool inventory, a JSON binding configuration maps each action to a
server and tool with optional output extractors, and a runtime
orchestration middleware executes the plan, accumulating extracted
outputs in an execution context and substituting them into later actions
via opaque \verb|${context.X}| templates. The architecture is implemented (the supporting software is available at
\url{https://github.com/PCfVW/hplan26-artifact}) and is demonstrated end-to-end from a
browser-based plan controller that streams per-action progress via
Server-Sent Events.

\paragraph{Paper scope.} This short paper reports a working coordination
architecture together with an end-to-end demonstration of feasibility on a
live multi-server biomedical pipeline; the contribution is the architecture
and its implementation rather than a benchmark comparison. It addresses the
workshop topics
\emph{applications of hierarchical planning}, \emph{techniques for
verifying solutions of hierarchical planning problems}, and \emph{using
Generative AI for hierarchical planning}---with the latter framed
inversely: HTN planning constrains LLM tool sequencing, rather than
LLMs guiding HTN search.

\paragraph{Organisation.} Section~\ref{sec:background} positions the work
against prior literature on classical HTN planning, planning-based
robotic orchestration, scientific workflow systems and LLM agents.
Section~\ref{sec:approach} presents the three-layer coordination
architecture and the architectural reduction it achieves over LLM-driven
sequencing. Section~\ref{sec:case-study} reports an end-to-end
live-server demonstration on a drug-target discovery workflow involving
eight third-party MCP servers querying real biological databases, with
supporting evidence from four other HTN domains.
Section~\ref{sec:conclusion} discusses limitations, the
generalisation of the MCP-isolation/HTN-decomposition correspondence,
and future work.

\section{Background and Related Work}
\label{sec:background}

\paragraph{HTN planning.} HTN planning's foundational systems include
SHOP and SHOP2 \citep{nau1999shop,nau2003shop2}; GTPyhop
\citep{nau2021gtpyhop} is a goal-task-network successor in which actions
and methods are Python functions, and methods return flat tuples of
subtasks. A goal-task-network planner admits both \emph{tasks} (decomposed
by methods) and \emph{goals} (state conditions to achieve); our domains use
only its task-decomposition (HTN) fragment, so every top-level task is
compound and every method returns a totally-ordered subtask tuple. Recent
work on HTN learning includes HTN-MAKER
\citep{htnmaker2008} and CURRICULAMA \citep{curriculama2024}; ChatHTN
\citep{chathtn2025} interleaves LLM-approximate decomposition with
symbolic verification; L2HP \citep{puertamerino2025} generates HDDL from
natural language.

\paragraph{Planning-based orchestration.} ROSPlan
\citep{cashmore2015rosplan} embeds a PDDL planner inside the Robot
Operating System for robotic mission orchestration, and is the closest
architectural antecedent to the present work: it too treats the planner
as a coordination layer above heterogeneous executors. Scientific
workflow systems---Nextflow \citep{ditommaso2017nextflow} chief
among them---solve a related reproducibility problem at a different layer
of the stack: the workflow description is authored manually and the
system handles distributed execution. The architecture proposed here
sits between the two: the workflow is generated by a planner from
declarative goals, and execution is distributed across MCP servers
rather than across compute resources. The closest formal antecedent, SHOP2
for Web-service composition \citep{sirin2004htn}---MCP servers being one
more kind of Web service---may \emph{execute} information-providing services
during planning; we instead keep planning and execution separate, fixing a
verifiable, inspectable plan before any server is contacted.

\paragraph{LLMs for planning and tool use.} ReAct \citep{yao2023react}
and Toolformer \citep{schick2023toolformer} are representative of the
LLM-driven tool-orchestration paradigm we contrast with. Two plan-once
variants are the natural baselines for our $N{+}1\to2$ argument: ReWOO
\citep{xu2023rewoo} substitutes variables into a pre-committed plan---close
to our \verb|${context.X}| bindings---and LLMCompiler
\citep{kim2024llmcompiler} executes an LLM-emitted task DAG with minimal
further calls. Both still synthesise the plan with an LLM, so it is neither
deterministic nor a verifiable symbolic artifact, whereas our HTN planner's
is; emitting tool-calling code in one pass \citep{felendler2026tool}
collapses the round-trips similarly but regenerates the orchestration logic
per request rather than yielding a replayable plan. A critical
strand \citep{valmeekam2023planning} argues that LLMs are unreliable
planners and should serve as translators or critics of formal plans
rather than generate them; LLM+P \citep{liu2023llmp} embodies that view
by having an LLM call a classical planner. A recent survey
\citep{huang2024understanding} maps the broader LLM-agent planning
landscape. Closer to our setting, work on LLM-generated planning models
\citep{oswald2024llm,guan2023leveraging} addresses the upstream
model-elicitation problem---complementary to ours rather than competing.

\section{The Three-Layer Coordination Architecture}
\label{sec:approach}

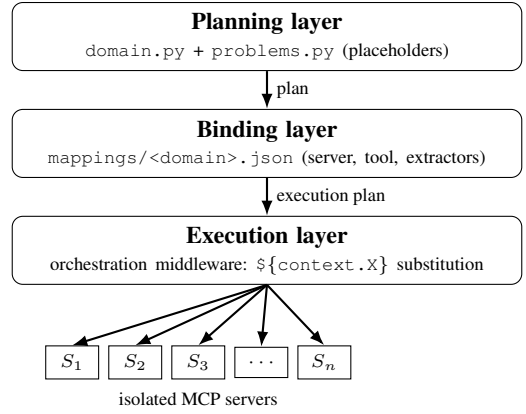
\begin{figure}[t]
  \centering
  \begin{tikzpicture}[
      layer/.style={rectangle, draw, rounded corners, minimum height=9mm,
                    text width=66mm, align=center, font=\small,
                    inner sep=2pt},
      server/.style={rectangle, draw, minimum height=4mm, minimum width=7mm,
                     font=\scriptsize},
      arrow/.style={-{Latex[length=2mm]}, thick},
      node distance=5mm
    ]
    \node[layer] (planning) {\textbf{Planning layer} \\
        \scriptsize\texttt{domain.py} + \texttt{problems.py}
        (placeholders)};
    \node[layer, below=of planning] (binding) {\textbf{Binding layer} \\
        \scriptsize\texttt{mappings/<domain>.json}
        (server, tool, extractors)};
    \node[layer, below=of binding] (execution) {\textbf{Execution layer} \\
        \scriptsize orchestration middleware:
        \texttt{\$\{context.X\}} substitution};

    \draw[arrow] (planning) -- node[right, font=\scriptsize]
        {plan} (binding);
    \draw[arrow] (binding) -- node[right, font=\scriptsize]
        {execution plan} (execution);

    \node[server, below=8mm of execution.south west, xshift=8mm] (s1) {$S_1$};
    \node[server, right=1.2mm of s1] (s2) {$S_2$};
    \node[server, right=1.2mm of s2] (s3) {$S_3$};
    \node[server, right=1.2mm of s3] (sd) {$\cdots$};
    \node[server, right=1.2mm of sd] (sn) {$S_n$};

    \draw[arrow] (execution.south) -- (s1.north);
    \draw[arrow] (execution.south) -- (s2.north);
    \draw[arrow] (execution.south) -- (s3.north);
    \draw[arrow] (execution.south) -- (sd.north);
    \draw[arrow] (execution.south) -- (sn.north);

    \node[below=0.5mm of s3, font=\scriptsize] {isolated MCP servers};
  \end{tikzpicture}
  \caption{Three-layer coordination architecture. The planning layer
    emits a sequence of primitive actions over placeholders; the
    binding layer maps each action to a (server, tool) pair and
    declares JSON-path output extractors; the execution layer walks
    the plan, captures extracted outputs into an execution context,
    and substitutes \texttt{\$\{context.X\}} templates as it goes.
    Each compound task in the planning layer decomposes into
    server-local primitives, mirroring MCP's server-isolation
    constraint.}
  \label{fig:architecture}
\end{figure}

\paragraph{Three-layer artifact decomposition.}
An executable MCP-orchestration domain consists of three artifacts at
three different abstraction layers (Figure~\ref{fig:architecture}),
separated for the same reasons a compiler separates source code,
linker symbol tables and runtime loader state.

(1)~\textbf{Planning layer:} a GTPyhop \texttt{domain.py} (actions and
methods) and \texttt{problems.py} (initial states and tasks). Action
arguments and state values are \emph{placeholders}---for example,
\verb|f"disease_id_for_{query}"|---so that the planner can verify task
ordering and precondition satisfiability without issuing real API calls.

(2)~\textbf{Binding layer:} a JSON configuration
\texttt{mappings/<domain>.json} that maps each HTN action to a (server,
tool) pair, optionally specifies a parameter-mapping table, and declares
JSON-path \emph{output extractors} (e.g.\
\texttt{disease\_id: data.search.hits[0].id}) that name run-time outputs.

(3)~\textbf{Execution layer:} a runtime orchestrator that walks the
plan, invokes MCP tools, applies the output extractors to capture named
outputs into an \emph{execution context}, and substitutes
\verb|${context.X}| template strings with their captured values when
later actions refer to them.

This separation is what makes the architecture verifiable: layer~(1) is
plannable in isolation, layer~(2) is schema-validated independently of
layer~(1), and layer~(3) operates on a purely declarative input
(plan~+~mapping). A reviewer or domain expert may inspect the planning
output before committing to physical execution.

\paragraph{Two orchestration approaches.}
We contrast two host orchestration strategies for the same multi-server
goal. \textbf{Approach~A} is LLM-driven: the LLM is given the tool
inventory and decides each next tool call from the current observation,
in the ReAct manner. An $N$-step workflow then requires $N{+}1$ LLM
inferences (one for the initial decision and one after each
observation). \textbf{Approach~B} is HTN-guided: the LLM is given the
goal once, calls the planner (\texttt{python\_find\_plan}), then the
execution tool (\texttt{python\_execute\_plan}), and the returned plan
is executed by the middleware without further LLM involvement. The
architectural reduction is exact: $N{+}1\to 2$ LLM inferences, a
constant independent of $N$ and of the per-inference latency. We report
only this architectural reduction, not a wall-clock speedup, as a
conservative claim (see Limitations).

\paragraph{The MCP-isolation/HTN-decomposition correspondence.}
A consequence of the three-layer separation is that MCP's
server-isolation constraint maps directly onto HTN's
compound-to-primitive decomposition: each primitive action is bound to a
single MCP server, so no method body ever expresses a cross-server action,
and cross-server data flow is materialised only at the binding/execution
boundary by the \verb|${context.X}|/JSON-path mechanism. HTN methods thus
need not know which servers their primitives target, and execution never
revisits planning.

\paragraph{Implementation.}
The planner is GTPyhop~\citep{nau2021gtpyhop}, in its maintained
v2.0.1 release \citep{gtpyhop201}; it is invoked from the host as an
MCP tool (\texttt{python\_find\_plan}) that consumes a domain
identifier and an initial state and returns a sequence of primitive
actions. A second tool (\texttt{python\_execute\_plan}) retrieves the
stored plan, joins it with the binding-layer JSON, and returns an
execution-plan response carrying a flag (\texttt{requires\_client\_orchestration:
true}) and the per-step (server, tool, arguments, extractors) tuples.
The orchestration middleware on the host detects this flag and walks
the plan, calling each MCP server in turn, accumulating extracted
outputs into the execution context, and substituting templates as it
goes. A browser-based plan controller exposes the same workflow over a
FastAPI backend with Server-Sent Events streaming for per-action
progress, enabling auditable runtime observation.

\section{Case Study: Drug-Target Discovery on Eight Live Servers}
\label{sec:case-study}

\paragraph{Setting.}
The drug-target-discovery domain coordinates eight third-party MCP
servers, open-source at the versions used, from the Augmented Nature
ecosystem~\citep{augmentednature}---OpenTargets, UniProt, Reactome,
KEGG, RCSB PDB, AlphaFold DB, ChEMBL and PubMed (NCBI E-utilities)---each
exposing a curated subset of its host database's REST or GraphQL
interface as MCP tools. The eight servers cover complementary
categories of evidence that arise in drug-target discovery:
disease--target associations, protein characterisation, mechanism,
structure, chemistry and literature
(Table~\ref{tab:tool-surface}). In aggregate they advertise 140
tools via MCP's \texttt{ListTools} contract; the demonstrated
workflow exercises 8 of them, one per plan step. None of
the servers requires API keys or authentication; all eight are
launched as Node.js stdio subprocesses and reused across runs. The
HTN domain comprises ten primitive actions (eight used in the
demonstrated plan) and six methods covering disease search, target enumeration, protein
characterisation, pathway analysis, structure retrieval, compound
identification and literature search. The binding-layer JSON
declares output extractors that materialise a typed dataflow graph
between adjacent steps: \texttt{disease\_id} from OpenTargets,
\texttt{first\_gene} from the disease-targets summary,
\texttt{uniprot\_accession} from UniProt, and so on through
\texttt{pmids} from PubMed.

\paragraph{Tool--action correspondence.}
The architecture commits to a one-to-one correspondence between
MCP tools and HTN primitive actions: each primitive action wraps
exactly one advertised tool. A graph database records the tools a
domain uses as nodes keyed by host server
and tool name, each linked to its primitive action by a mapping edge
carrying the parameter bindings and JSON-path output extractors;
these are hand-written, since none of the eight servers declares an
output schema. Authoring an HTN domain for a specific protocol then
reduces to (i)~selecting the tools the protocol invokes and
(ii)~composing their actions into HTN methods that encode the
protocol's procedural decomposition. The companion
paper~\citep{anonymous2026authoring} reports an LLM-driven
pipeline that performs both steps: given a user-supplied protocol
description, it generates a \texttt{domain.py} (the retrieved
actions, plus methods encoding the protocol) and a
\texttt{problems.py} (initial states and tasks) that fit the
binding-layer schema used here.

\begin{table}[t]
  \centering
  \begin{small}
  \setlength{\tabcolsep}{4.5pt}
  \begin{tabular}{@{}llr@{}}
    \toprule
    Evidence category & Server(s) & Tools \\
    \midrule
    Disease--target associations           & OpenTargets       & 6        \\
    Protein characterisation               & UniProt           & 26       \\
    Mechanism (pathways)                   & Reactome, KEGG    & 8\,+\,33 \\
    Structure (experimental, predicted)    & PDB, AlphaFold    & 5\,+\,19 \\
    Chemistry (compounds, bioactivity)     & ChEMBL            & 27       \\
    Literature (cross-cutting)             & PubMed            & 16       \\
    \midrule
    Total advertised; exercised by demo    &                   & 140; 8   \\
    \bottomrule
  \end{tabular}
  \end{small}
  \caption{Tool surface of the eight Augmented Nature MCP servers,
    grouped into six evidence categories of drug-target discovery.
    The live demonstration of Table~\ref{tab:live-runs} exercises 8
    of the 140 advertised tools.}
  \label{tab:tool-surface}
\end{table}

\paragraph{End-to-end execution from a browser.}
We exercised the full architecture by clicking the plan controller's
\emph{Execute} action on a stored breast-cancer plan three times in
succession. The FastAPI backend received a
\texttt{POST /api/execute/start} with \texttt{mock=false}, which routed
to the real-server execution path and connected the eight stdio
subprocesses (subprocesses are spawned at backend startup and reused
across runs). Each run streamed per-step events back to the browser
via Server-Sent Events. Table~\ref{tab:live-runs} reports the three
runs, each on the same plan key, each producing the same eight-step
sequence in the same server order, each returning real biological
data from live APIs.

\begin{table}[t]
  \centering
  \begin{small}
  \begin{tabular}{@{}clrr@{}}
    \toprule
    Run & Subprocesses    & Wall-clock & Executed \\
    \midrule
    1   & cold (just spawned) & 35\,s  & 8/8   \\
    2   & warm (reused)       & 29\,s  & 8/8   \\
    3   & warm (reused)       & 15\,s  & 8/8   \\
    \bottomrule
  \end{tabular}
  \end{small}
  \caption{Three replays of the same eight-step drug-target-discovery
    plan against eight live Augmented Nature MCP servers, executed
    from a browser-based plan controller. Each run executes the same
    sequence (\texttt{search\_diseases} on OpenTargets;
    \texttt{get\_disease\_targets\_summary} on OpenTargets;
    \texttt{search\_by\_gene} on UniProt;
    \texttt{find\_pathways\_by\_gene} on Reactome;
    \texttt{search\_by\_uniprot} on PDB;
    \texttt{get\_structure} on AlphaFold;
    \texttt{search\_by\_uniprot} on ChEMBL;
    \texttt{search\_articles} on PubMed) with identical step ordering
    and identical server distribution. Wall-clock is dominated by
    slow external APIs whose latency varies across sessions (ChEMBL
    here, PubMed in others).}
  \label{tab:live-runs}
\end{table}

The point of Table~\ref{tab:live-runs} is not the wall-clock numbers,
which are dominated by external API latency, but the structural
reproducibility: the same plan key produces the same step sequence
and the same server distribution on every run, with identical
\verb|${context.X}| substitution chains
(\texttt{disease\_id}~$\to$ \texttt{first\_gene}~$\to$
\texttt{uniprot\_accession}~$\to$ \ldots).
This is the property that LLM-driven orchestration cannot guarantee.

\paragraph{Scaling: bio-laboratory PCR.}
A second case study, on a simulated three-server PCR-preparation domain
spanning movement, liquid-handling and module-control servers, shows
that the architecture scales: HTN plans range from 55 primitive actions
for a 4-sample PCR to 611 for a 96-well plate (plan length following the
closed form $31 + 6n + 2(\lceil n/40 \rceil - 1)$, e.g.\ $n{=}4 \to 55$ and
$n{=}96 \to 611$), executed at a
steady ${\sim}$16.5 actions/s on simulated MCP clients with 100\%
success across all configurations. The simulation faithfully implements
the Opentrons Flex API surface; a hardware deployment would substitute
physical servers without changing the planning or binding layers.

\paragraph{Coverage.}
Three further HTN domains have been authored against this
architecture: a four-server DNA-extraction protocol on the Opentrons
Flex (Omega HDQ; 89--129 actions across three scenarios), a multiscale
TNF cancer-modelling workflow joining MaBoSS and
PhysiCell~\citep{ruscone2025biomodelling,ruscone2025prototyping},
and a three-server pick-and-place robotics scenario. Of these
three, only the DNA-extraction domain currently has a complete
binding-layer mapping; the other two exercise the planning layer
alone. We treat this as evidence that the authoring problem (covered
in the companion paper \citep{anonymous2026authoring}) and the
coordination problem (covered here) are independently tractable.

\section{Discussion and Conclusion}
\label{sec:conclusion}

\paragraph{Reproducibility and auditability.}
The same plan key replayed against the eight live Augmented Nature
servers produced the same step sequence on three successive runs;
intermediate values (extracted disease IDs, gene names, UniProt
accessions) varied as little as the underlying APIs varied, but the
plan structure itself was invariant under replay. The Server-Sent
Events stream provides a per-step audit trail---start, complete, error,
with millisecond durations---that a domain expert can inspect post hoc,
in contrast to an LLM agent's reasoning trace, latent in
token logs.

\paragraph{Limitations.}
We do not report a measured LLM-driven baseline; the
$N{+}1\to 2$ inference reduction follows from the architecture and
is exact, but a wall-clock speedup claim would require an empirical
LLM tool-use comparison that we leave to future work. The architecture's
default execution mode is mocked, by design, so that demonstration
runs do not inadvertently consume third-party APIs; live-server
execution is opt-in. We adopt MCP because it standardises tool discovery
(\texttt{ListTools}) across heterogeneous providers, not because it is
uniquely suited---the architecture is interface-agnostic over REST, GraphQL
or CLI tools under the same isolation assumption. The planner selects
methods by precondition, so domains may branch, though the demonstrated
drug-discovery domain is largely linear; richer branching is future work.
The companion paper's evaluation
\citep{anonymous2026authoring} examines the upstream authoring problem
and reports per-pass output-quality measurements on a separate
fixture; the two papers can be read independently.

\paragraph{Generalisation.}
The MCP-isolation/HTN-decomposition correspondence is an instance of a
broader pattern: any host-mediated multi-tool ecosystem with isolated
tool providers admits a hierarchical-planning coordination layer in
which compound tasks decompose into provider-local primitives. We
conjecture the same architecture applies, with modest changes, to
function-calling APIs over isolated REST endpoints, to multi-agent
systems where agents expose tools but cannot directly delegate, and
to scientific workflow managers that internalise tool selection.

\paragraph{Conclusion.}
All three problems with LLM-driven cross-server orchestration---cost
($N{+}1$ inferences), reproducibility and inspectability before
execution---admit one architecturally clean fix: separate planning from
binding from execution, emit the plan once, and let a runtime middleware
execute it deterministically, so the planning layer's output is a
verifiable artifact independent of execution. We have demonstrated this
end-to-end on eight live third-party servers from a browser-based
controller; the supporting code is available at
\url{https://github.com/PCfVW/hplan26-artifact} and archived as
\url{https://doi.org/10.5281/zenodo.22999239}.



\bibliography{htn-coordination-mcp}

\end{document}